\documentclass[a4paper]{article}
\pdfoutput=1

\usepackage{INTERSPEECH2019}

\usepackage[utf8]{inputenc}

\usepackage{amsmath}
\usepackage{graphicx}
\usepackage{booktabs}
\usepackage{hyperref}
\usepackage{url}

\usepackage{paralist}
\usepackage[inline]{enumitem}
\usepackage{listings}
\usepackage{microtype}
\usepackage{multirow}
\usepackage{cite}

\usepackage{todonotes}

\usepackage{tikz}
\usetikzlibrary{positioning, shapes, fit, arrows}
\usetikzlibrary{external}

\newcommand{\cascademodel}[1]{ST#1$\,\rightarrow\,$TTS cascade}

\newcommand{\modelname}{Translatotron}

\title{Direct speech-to-speech translation with a sequence-to-sequence model}

\name{Ye Jia\textsuperscript{*}\thanks{* Equal contribution.}, Ron J. Weiss\textsuperscript{*}, Fadi Biadsy, Wolfgang Macherey, Melvin Johnson, Zhifeng Chen, Yonghui Wu}
\address{Google}
\email{\texttt{\{jiaye,ronw\}@google.com}}

\begin{document}

\maketitle
\begin{abstract}
  We present an attention-based sequence-to-sequence neural network which can directly translate speech from one language into speech in another language, without relying on an intermediate text representation.
  The network is trained end-to-end, learning to map speech spectrograms into target spectrograms in another language, corresponding to the translated content (in a different canonical voice).
  We further demonstrate the ability to synthesize translated speech using the voice of the source speaker.   We conduct experiments on two Spanish-to-English speech translation datasets,
  and find that the proposed model slightly underperforms a baseline cascade of a direct speech-to-text translation model and a text-to-speech synthesis model, demonstrating the feasibility of the approach on this very challenging task.

\end{abstract}


\section{Introduction}
\label{sec:intro}

We address the task of speech-to-speech translation (S2ST): translating speech in one language into speech in another.
This application is highly beneficial for breaking down communication barriers between people who do not share a common language.
Specifically, we investigate whether it is possible to train model to accomplish this task directly, without relying on an intermediate text representation.
This is in contrast to conventional S2ST systems which are often broken down into three components: automatic speech recognition (ASR), text-to-text machine translation (MT), and text-to-speech (TTS) synthesis \cite{lavie1997janus, wahlster2000verbmobil, nakamura2006atr,itu-f745}.

Cascaded systems have the potential problem of errors compounding between components, e.g.\  recognition errors leading to larger translation errors.
Direct S2ST models avoid this issue by training to solve the task end-to-end.
They also have advantages over cascaded systems in terms of
reduced computational requirements and lower inference latency since only one decoding step is necessary, instead of three. In addition, direct models are naturally capable of retaining paralinguistic and non-linguistic information during translation, e.g.\ maintaining the source speaker's voice, emotion, and prosody, in the synthesized translated speech.
Finally, directly conditioning on the input speech makes it easy to learn to generate fluent pronunciations of words which do not need to be translated, such as names. 

However, solving the direct S2ST task is especially challenging for several reasons.
Fully-supervised end-to-end training
requires
collecting a large set of input/output speech pairs.
Such data are more difficult to collect compared to parallel text pairs for MT, or speech-text pairs for ASR or TTS.
Decomposing into smaller tasks can take advantage of the lower training data requirements compared to a monolithic speech-to-speech model, and can result in a more robust system for a given training budget.
Uncertain alignment between two spectrograms whose underlying spoken content differs also poses a major training challenge. 

In this paper we demonstrate \emph{\modelname{}}\footnote{Audio samples are available at \begin{scriptsize}
\url{https://google-research.github.io/lingvo-lab/translatotron}
\end{scriptsize}.
}, a 
direct speech-to-speech translation model which is trained end-to-end.
To facilitate training without predefined alignments, we leverage high level representations of the source or target content in the form of transcriptions,
essentially multitask training with speech-to-text tasks.
However no intermediate text representation is used during inference. The model does not perform as well as a baseline cascaded system.
Nevertheless, it demonstrates a proof of concept and serves as a starting point for future research.

Extensive research has studied methods for combining different sub-systems within cascaded speech translation systems.
\cite{ney1999speech, matusov2005integration} gave MT access to the lattice of the ASR. 
\cite{vidal1997finite, casacuberta2004some} integrated acoustic and translation models using a stochastic finite-state transducer which can
decode the translated text directly using Viterbi search.
For synthesis,
\cite{aguero2006prosody} used unsupervised
clustering to find F0-based prosody features and transfer intonation from source speech and target.
\cite{do2017toward} augmented MT to jointly predict translated words and emphasis, in order to improve expressiveness of the synthesized speech.
\cite{kano2018end} used a neural network to transfer duration and power from the source speech to the target.
\cite{kurimo2010personalising} transfered source speaker's voice to the synthesized translated speech by mapping hidden Markov model states from ASR to TTS.
Similarly, recent work on neural TTS has focused on adapting to new voices with limited reference data \cite{nachmani2018fitting,arik2018neural,jia2018transfer,chen2018sample}.

Initial approaches to end-to-end speech-to-text translation (ST) \cite{berard2016listen, berard2018end} performed worse than a cascade of an ASR model and an MT model.
\cite{weiss2017sequence, anastasopoulos2018tied} achieved better end-to-end performance by leveraging weakly supervised data with multitask learning.
\cite{jia2018leveraging} further showed that use of synthetic training data can work better than multitask training.
In this work we take advantage of both synthetic training targets and multitask training.

The proposed model resembles recent sequence-to-sequence models for voice conversion,
the task of recreating an utterance in another person's voice
\cite{haque2018conditional, zhang2019sequence, biadsy2019parrotron}. For example, \cite{zhang2019sequence} proposes an attention-based model to generate spectrograms in the target voice based on input features (spectrogram concatenated with ASR bottleneck features) from the source voice.
In contrast to S2ST, the input-output alignment for voice conversion is simpler and approximately monotonic.
\cite{zhang2019sequence} also trains models that are specific to each input-output speaker pair (i.e.\ one-to-one conversion), whereas we explore many-to-one and many-to-many speaker configurations. Finally, \cite{guo2019end} demonstrated an attention-based direct S2ST model on a toy dataset with a 100 word vocabulary.
In this work we train on real speech, including spontaneous telephone conversations, at a much larger scale.

 \section{Speech-to-speech translation model}
\label{sec:model}

\begin{figure}[t]
  \centering

\scalebox{0.81}{
\begin{tikzpicture}[auto, font=\small\sffamily, node distance=2.0cm,auto,>=latex']

  \pgfdeclarelayer{back}
  \pgfdeclarelayer{middle}
  \pgfsetlayers{back,middle,main}

  \tikzstyle{optional_label} = [text=darkgray]
  \tikzstyle{optional} = [draw=darkgray]
  \tikzstyle{cell} = [rectangle, draw, fill=blue!20, text width=4.2em, text centered, minimum height=4.2ex]
  \tikzstyle{enccell} = [cell, text width=4.0em]
  \tikzstyle{auxcell} = [cell, optional, optional_label, text width=, fill=blue!10]
  \tikzstyle{attnwidth} = [text width=4.1em]
  \tikzstyle{component} = [draw, minimum height=4.5ex, minimum width=6em, fill=blue!20, rounded corners=4, inner ysep=3pt, inner xsep=6pt]
  \tikzstyle{component_label} = [inner sep=1pt, align=left, font=\scriptsize\sffamily]
  \tikzstyle{io_label} = [align=center, inner sep=2pt, font=\scriptsize\sffamily]
  \tikzstyle{point} = [inner sep=0pt, outer sep=0pt]

  \node [io_label, name=input, align=center] {log-mel spectrogram\\(Spanish)};

  \node[enccell, minimum height=2.2cm, above=0.3cm of input] (enc_stack) {8-layer Stacked BLSTM Encoder};
  \iffalse
  \node[enccell, above=0.3cm of input] (enc14) {4x BLSTM};
  \node[point, above=0.3cm of enc14] (encdots) {};
  \node[enccell, above=0.3cm of encdots] (enc56) {BLSTM};
  \node[enccell, above=0.4cm of enc56] (enc78) {BLSTM};
  \node [component_label, above=0.3cm of enc78] (enc_stack_label) {Encoder};
  \begin{pgfonlayer}{back}
    \node[component, fit=(enc14)(enc56)(enc78)(enc_stack_label), fill=gray!20] (enc_stack) {};
  \end{pgfonlayer}
  \fi

  \node[cell, above right=-0.1cm and 0.3cm of enc_stack, text width=, inner xsep=1pt] (concat) {concat};
  \node[cell, optional, optional_label, above=0.25cm of concat, inner xsep=1pt, fill=green!10] (enc_speaker) {Speaker Encoder};
  \node [io_label, optional_label, left=0.4cm of enc_speaker, inner xsep=0pt, align=center] (input_speaker) {speaker\\reference\\utterance};

  \node[auxcell, above right=-1.9cm and 0.5cm of enc_stack] (attn1) {Attention};
  \node[auxcell, above right=-1.1cm and 0.5cm of enc_stack] (attn2) {Attention};
  \node[cell, right=0.3cm of concat] (attn3) {Multihead Attention};
  \iffalse
  \node[auxcell, above right=-0.1cm and 0.5cm of enc14] (attn1) {Attention};
  \node[auxcell, above right=-0.15cm and 0.5cm of enc56] (attn2) {Attention};
  \node[cell, above right=-0.2cm and 0.5cm of enc_stack] (attn3) {Multihead Attention};
  \fi

  \node[auxcell, right=0.3cm of attn1] (dec1) {$2\times$ LSTM Decoder};
  \node[auxcell, right=0.3cm of attn2] (dec2) {$2\times$ LSTM Decoder};
  \node[cell, right=0.3cm of attn3, text width=5.6em, inner xsep=2pt] (dec3) {Spectrogram   Decoder};

  \node[cell, above right=0.3cm and -0.35cm of dec3, fill=red!20, text width=3.5em] (vocoder) {Vocoder};

  \node [io_label, optional_label, right=0.5cm of dec1] (output1) {phonemes\\(Spanish)};
  \node [io_label, optional_label, right=0.5cm of dec2] (output2) {phonemes\\(English)};
  \node [io_label, right=0.3cm of dec3] (output3) {  linear freq\\spectrogram\\(English)};
  \node [io_label, right=0.25cm of vocoder] (output4) {waveform\\(English)};
  
  \node [component_label, optional_label, below=0.15cm of dec1] (aux_stack_label) {Auxiliary recognition tasks};
  \begin{pgfonlayer}{back}
    \node[component, optional, fill=gray!10, fit=(attn1)(attn2)(dec1)(dec2)(aux_stack_label)(output1)(output2)] (aux_stack) {};
  \end{pgfonlayer}

  \draw[->]
    (input) edge (enc_stack)
    (enc_stack) |- (concat)
    (concat) edge (attn3);
  \iffalse
  \draw[->]
    (input) edge (enc14)
    (encdots) edge (enc56)
    (enc56) edge (enc78)
    (enc78) |- (attn3);
  \draw[->, dotted]
    (enc14) edge (encdots);
  \fi
  \draw[->]
    (attn3) edge (dec3)
    (dec3) -| (vocoder);
  \begin{pgfonlayer}{back}
    \draw[->, gray]
      (enc_stack) |- (attn1)
      (attn1) edge (dec1)
      (dec1) edge (output1);
    \draw[->, gray]
      (enc_stack) |- (attn2)
      (attn2) edge (dec2)
      (dec2) edge (output2);
  \end{pgfonlayer}

  \draw [->, gray] (input_speaker) edge (enc_speaker);
  \draw [->, gray] (enc_speaker) -- (concat);
  
  \draw [->] (vocoder) -- (output4);
  
\end{tikzpicture}
}
\vskip-1.3ex
\caption{Proposed model architecture,
which generates English speech (top right) from Spanish speech (bottom left), and an optional speaker reference utterance (top left) which is only used for voice transfer experiments in Section~\ref{sec:voice_transfer}.
The model is multitask trained to predict source and target phoneme transcripts as well, however these auxiliary tasks  are not used during inference.
Optional components are drawn in light colors.
}
\vskip-2ex
\label{fig:model}
\end{figure}
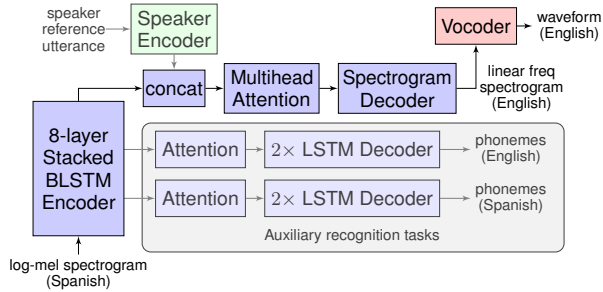

An overview of the proposed \modelname{} model architecture is shown in Figure~\ref{fig:model}.
Following \cite{jia2018transfer, shen2018natural}, it is composed of several separately trained components:
\begin{inparaenum}[1)]
\item an \emph{attention-based sequence-to-sequence network} (blue) which generates target spectrograms,
\item a \emph{vocoder} (red) which converts target spectrograms to time-domain waveforms,
and,
\item optionally, a pretrained \emph{speaker encoder} (green) which can be used to condition the decoder on the identity of the source speaker, enabling cross-language voice conversion \cite{machado2010voice} simultaneously with translation.
\end{inparaenum}

The sequence-to-sequence encoder stack maps 80-channel log-mel spectrogram input features into hidden states which are passed through an attention-based alignment mechanism to condition an autoregressive decoder, which predicts 1025-dim
log spectrogram frames corresponding to the translated speech. Two optional auxiliary decoders, each with their own attention components, predict source and target phoneme sequences.

Following recent speech translation \cite{jia2018leveraging} and recognition \cite{chiu2017state} models, the encoder is composed of a stack of 8 bidirectional LSTM layers.
As shown in Fig.~\ref{fig:model}, the final layer output is passed to the primary decoder, whereas intermediate activations are passed to auxiliary decoders predicting phoneme sequences.We hypothesize that early layers of the encoder are more likely to represent the source content well, while deeper layers might learn to encode more information about the target content.

The spectrogram decoder uses an architecture similar to Tacotron~2 TTS model \cite{shen2018natural}, including pre-net, autoregressive LSTM stack, and post-net components.
We make several changes to it in order to adapt to the more challenging S2ST task.
We use multi-head additive attention \cite{vaswani2017attention} with 4 heads instead of location-sensitive attention, which shows better performance in our experiments. We also use a significantly narrower 32 dimensional pre-net bottleneck compared to 256-dim in \cite{shen2018natural}, which we find to be critical in picking up attention during training. We also use reduction factor \cite{yx2017tacotron} of 2, i.e.\ predicting two spectrogram frames for each decoding step.  Finally, consistent with results on translation tasks \cite{wu2016google,weiss2017sequence}, we find that using a deeper decoder containing 4 or 6 LSTM layers leads to good performance.

We find that  multitask training is critical in solving the task, which we accomplish by integrating auxiliary decoder networks to predict phoneme
sequences corresponding to the source and/or target speech. Losses computed using these auxiliary recognition networks are used during training, which help the primary spectrogram decoder to learn attention. They are not used during inference.
In contrast to the primary decoder, the auxiliary decoders use 2-layer LSTMs with single-head additive attention \cite{bahdanau2014neural}.
All three decoders use attention dropout and LSTM zoneout regularization \cite{krueger2016zoneout}, all with probability 0.1.
Training uses the Adafactor optimizer \cite{shazeer2018adafactor} with a batch size of 1024.

Since we are only demonstrating a proof of concept,
we primarily rely on the low-complexity Griffin-Lim \cite{griffin1984signal} vocoder in our experiments. However, we use a WaveRNN \cite{kalchbrenner2018efficient} neural vocoder when evaluating speech naturalness 
in listening tests.

Finally, in order to control the output speaker identity we incorporate an optional speaker encoder network as in \cite{jia2018transfer}.
This network is discriminatively pretrained on a speaker verification task and is not updated during the training of \modelname{}.
We use the \textit{dvector V3} model from \cite{zhang2018fully}, trained on a larger set of
851K speakers across 8 languages including English and Spanish.
The model computes a 256-dim speaker embedding from the speaker reference utterance, which is passed into a linear projection layer (trained with the sequence-to-sequence model) to reduce the dimensionality to 16. This is critical to generalizing to source language speakers which are unseen during training. %
 \section{Experiments}
\label{sec:experiments}

We study two Spanish-to-English translation datasets: the large scale ``conversational'' corpus of parallel text and read speech pairs from \cite{jia2018leveraging}, and the Spanish Fisher corpus of telephone conversations and corresponding English translations \cite{post2013improved}, which is smaller and more challenging due to the spontaneous and informal speaking style.
In Sections~\ref{sec:proprietary}~and~\ref{sec:fisher}, we \emph{synthesize} target speech from the target transcript using a single (female) speaker English TTS system; In Section~\ref{sec:voice_transfer}, we use real human target speech for voice transfer experiments on the conversational dataset.
Models were implemented using the Lingvo framework \cite{shen2019lingvo}.
See Table~\ref{tbl:hyperparams} for
dataset-specific
hyperparameters.

\begin{table}[t]
\setlength{\tabcolsep}{0.5em}
\centering
\begin{footnotesize}\caption{Dataset-specific model hyperparameters.}
\vskip-1ex
\begin{tabular}{l@{\hspace{0.2em}}rr}
    \toprule
      & Conversational & Fisher \\
    \midrule
    Num train examples & 979k & 120k \\
    Input / output sample rate (Hz) & 16k / 24k & 8k / 24k \\
    Learning rate  & 0.002  & 0.006 \\
    Encoder BLSTM  & 8$\times$1024 & 8$\times$256 \\
    Decoder LSTM & 6$\times$1024 & 4$\times$1024 \\
    Auxiliary decoder LSTM & 2$\times$256 & 2$\times$256 \\
    \hspace{1em} source / target input layer & 8 / 8 & 4 / 6 \\
    \hspace{1em} dropout prob & 0.2 & 0.3 \\
    \hspace{1em} loss decay & constant 1.0 &  
    $0.3\rightarrow0.001$\\&& at 160k steps\\
    Gaussian weight noise stddev & none & 0.05 \\
    \bottomrule
\end{tabular}
\label{tbl:hyperparams}
\end{footnotesize}\vskip-2ex
\end{table}

To evaluate speech-to-speech translation performance
we compute BLEU scores \cite{papineni-EtAl:2002:ACL} as an objective measure of speech intelligibility and translation quality, by using a pretrained ASR system to recognize the generated speech, and comparing the resulting transcripts to ground truth reference translations.
Due to potential recognition errors (see Figure~\ref{fig:example_outputs-fisher}), this can be thought of as a lower bound on the underlying translation quality.
We use the \textit{16k Word-Piece} attention-based ASR model from \cite{irie2019model} trained on the 960 hour LibriSpeech corpus \cite{panayotov2015librispeech}, which obtained word error rates of 4.7\% and 13.4\% on the test-clean and test-other sets, respectively. 
In addition, we conduct listening tests to measure subjective speech naturalness mean opinion score (MOS), as well as speaker similarity MOS for voice transfer.

\subsection{Conversational Spanish-to-English}
\label{sec:proprietary}

This proprietary dataset described in \cite{jia2018leveraging}
was obtained by crowdsourcing humans to read the both sides of a conversational Spanish-English MT dataset.
In this section, instead of using the human target speech, we use a TTS model to synthesize target speech in a single female English speaker's voice in order to simplify the learning objective.
We use an English Tacotron~2 TTS model~\cite{shen2018natural} but use a Griffin-Lim vocoder for expediency. In addition, we augment the input source speech by adding background noise and reverberation in the same manner as \cite{jia2018leveraging}.

The resulting dataset contains 979k parallel utterance pairs, containing 1.4k hours of source speech and 619 hours of synthesized target speech. The total target speech duration is much smaller because the TTS output is better endpointed, and contains fewer pauses. 9.6k pairs are held out for testing.

Input feature frames are created by stacking 3 adjacent frames of an 80-channel log-mel spectrogram as in \cite{jia2018leveraging}.
The speaker encoder was not used in these experiments since the target speech always came from the same speaker.

\begin{table}[t]
\centering
\setlength{\tabcolsep}{0.5em}
\begin{small}
\caption{Conversational test set performance. Single reference BLEU
and Phoneme Error Rate (PER) of aux decoder outputs.}
\vskip-1ex
\begin{tabular}{l@{\hspace{0.5em}}rrr}
    \toprule
    Auxiliary loss & BLEU & Source PER & Target PER \\
    \midrule
    None            &  0.4 &   - &    - \\
    Source          & 42.2 & 5.0 &    - \\
    Target          & 42.6 &   - & 20.9 \\
    Source + Target & 42.7 & 5.1 & 20.8 \\
    \midrule
    \cascademodel{ \cite{jia2018leveraging}}
                    & 48.7 &   - &   - \\
    
    Ground truth    & 74.7 &   - &   - \\
    \bottomrule
\end{tabular}
\label{tbl:proprietary}
\end{small}
\vskip-2ex
\end{table}

Table~\ref{tbl:proprietary} shows performance of the model trained using different combinations of auxiliary losses, compared to a baseline \cascademodel{} model using a speech-to-text translation model \cite{jia2018leveraging} trained on the same data, and the same Tacotron~2 TTS model used to synthesize training targets. Note that the ground truth BLEU score is below 100 due to ASR errors during evaluation, or TTS failure when synthesizing the ground truth.

Training without auxiliary losses leads to extremely poor performance.   
The model correctly synthesizes common words and simple phrases, e.g.\ translating ``hola'' to ``hello''.
However, it does not consistently translate full utterances.
While it always generates plausible speech sounds in the target voice, the output can be independent of the input, composed of a string of nonsense syllables.
This is consistent with failure to learn to attend to the input, and reflects the difficulty of the direct S2ST task.

Integrating auxiliary phoneme
recognition tasks helped regularize the encoder and enabled the model to
learn attention,
dramatically improving performance. The target phoneme PER is much higher than on source phonemes, reflecting the difficulty of the corresponding translation task.
Training using both auxiliary tasks achieved the best quality, but the performance difference between different combinations is small.
Overall, there remains a gap of 6 BLEU points to the baseline, indicating room for improvement.
Nevertheless, the relatively narrow gap demonstrates the potential of the end-to-end approach. 

\subsection{Fisher Spanish-to-English}
\label{sec:fisher}

This dataset contains about 120k parallel utterance pairs\footnote{This is a subset of the Fisher data due to TTS errors on target text.}, spanning 127 hours of source speech. Target speech is synthesized using Parallel WaveNet~\cite{pmlr-v80-oord18a} using the same voice as the previous section.
The result contains 96 hours of synthetic target speech.

Following \cite{weiss2017sequence}, input features were constructed by stacking 80-channel log-mel spectrograms, with deltas and accelerations.
Given the small size of the dataset compared to that in Sec.~\ref{sec:proprietary},
we found that obtaining good performance required significantly more careful regularization and tuning.
As shown in Table~\ref{tbl:hyperparams}, we used narrower encoder dimension of 256, a shallower 4-layer decoder, and added Gaussian weight noise to all LSTM weights as regularization, as in \cite{weiss2017sequence}.
The model was especially sensitive to the auxiliary decoder hyperparameters, with the best performance coming when passing activations from intermediate layers of the encoder stack as inputs to the auxiliary decoders,
using slightly more aggressive dropout of 0.3, and decaying the auxiliary loss weight over the course of training in order to encourage the model training to fit the primary S2ST task.

\begin{table}[t]
\centering
\setlength{\tabcolsep}{1.25ex}
\begin{small}
\caption{Performance on the Fisher Spanish-English task in terms of 4-reference BLEU score on 3 eval sets.}
\vskip-1ex
\begin{tabular}{l@{\hspace{1.0em}}rrr}
    \toprule
    Auxiliary loss  & dev1 & dev2 & test \\
    \midrule
    None            &  0.4 &  0.6 &  0.6 \\
    Source          &  7.4 &  8.0 &  7.2 \\
    Target          & 20.2 & 21.4 & 20.8 \\
    Source + Target & 24.8 & 26.5 & 25.6 \\
    \addlinespace
    Source + Target (1-head attention)
                    & 23.0 & 24.2 & 23.4 \\
    Source + Target (encoder pre-training)
                    & 30.1 & 31.5 & 31.1 \\
    \midrule
    \cascademodel{ \cite{weiss2017sequence}} & 39.4 & 41.2 & 41.4 \\
    Ground truth    & 82.8 & 83.8 & 85.3 \\
    \bottomrule
\end{tabular}
\label{tbl:fisher}
\end{small}
\vskip-0.5ex
\end{table}

Experiment results are shown in Table~\ref{tbl:fisher}.
Once again using two auxiliary losses works best, but in contrast to Section~\ref{sec:proprietary}, there is a large performance boost relative to using either one alone.
Performance using only the source recognition loss is very poor, indicating that learning alignment on this task is especially difficult without strong supervision on the translation task.

We found that 4-head attention works better than one head, unlike the conversational task, where both attention mechanisms had similar performance.
Finally, as in \cite{jia2018leveraging}, we find that pre-training the bottom 6 encoder layers on an ST task improves BLEU scores by over 5 points.
This is the best performing direct S2ST model, obtaining 76\% of the baseline performance. 

\subsection{Subjective evaluation of speech naturalness}

To evaluate synthesis quality of the best performing models from Tables~\ref{tbl:proprietary}~and~\ref{tbl:fisher} we use the framework from \cite{jia2018transfer} to crowdsource 5-point MOS evaluations based on subjective listening tests.
1k examples were rated for each dataset, each one by a single rater.
Although this evaluation is expected to be independent of the correctness of the translation, translation errors can result in low scores for examples raters describe as ``not understandable''.

\begin{table}[t]
\centering
\begin{small}
\caption{Naturalness MOS with 95\% confidence intervals.
The ground truth for both datasets are synthetic English speech. }
\vskip-1ex
\setlength{\tabcolsep}{0.25em}
\begin{tabular}{l@{\hspace{0.7em}}l@{\hspace{0.25em}}cc}
    \toprule
    Model           & Vocoder & Conversational & Fisher-test \\
    \midrule
    \modelname{}    & WaveRNN & $4.08 \pm 0.06$ & $3.69 \pm 0.07$ \\
                    & Griffin-Lim & $3.20 \pm 0.06$ & $3.05 \pm 0.08$ \\
    \addlinespace
    ST$\rightarrow$TTS                     & WaveRNN & $4.32 \pm 0.05$ & $4.09 \pm 0.06$ \\
                    & Griffin-Lim & $3.46 \pm 0.07$ & $3.24 \pm 0.07$ \\
    \midrule
    Ground truth    & Griffin-Lim & $3.71 \pm 0.06$ & - \\
                    & Parallel WaveNet & - & $3.96 \pm 0.06$ \\
    \bottomrule
\end{tabular}
\label{tbl:naturalness}
\end{small}
\vskip-2ex
\end{table}

Results are shown in Table~\ref{tbl:naturalness}, comparing different vocoders where results with Griffin-Lim correspond to identical model configurations as Sections~\ref{sec:proprietary}~and~\ref{sec:fisher}.
As expected, using WaveRNN vocoders dramatically improves ratings over Griffin-Lim into the ``Very Good'' range (above 4.0).
Note that it is most fair to compare the Griffin-Lim results to the ground truth training targets since they were generated using corresponding lower quality vocoders. 
In such a comparison it is clear that the S2ST models do not score as highly as the ground truth.

\iffalse  \begin{figure}[t]
  \centering
  \vskip-1ex
  \includegraphics[width=\columnwidth]{figs/interspeech2019-translatotron-conversational-10025411426812728167}
   \vskip-2.5ex
  \caption{Example mel spectrograms from the Conversational dataset, with corresponding ASR transcripts used when computing BLEU scores.
       Audio is available on the companion webpage.
       }
  \label{fig:example_outputs-conversational}
\end{figure}
\fi

\begin{figure}[t]
  \centering
  \vskip-1ex
  \includegraphics[width=\columnwidth]{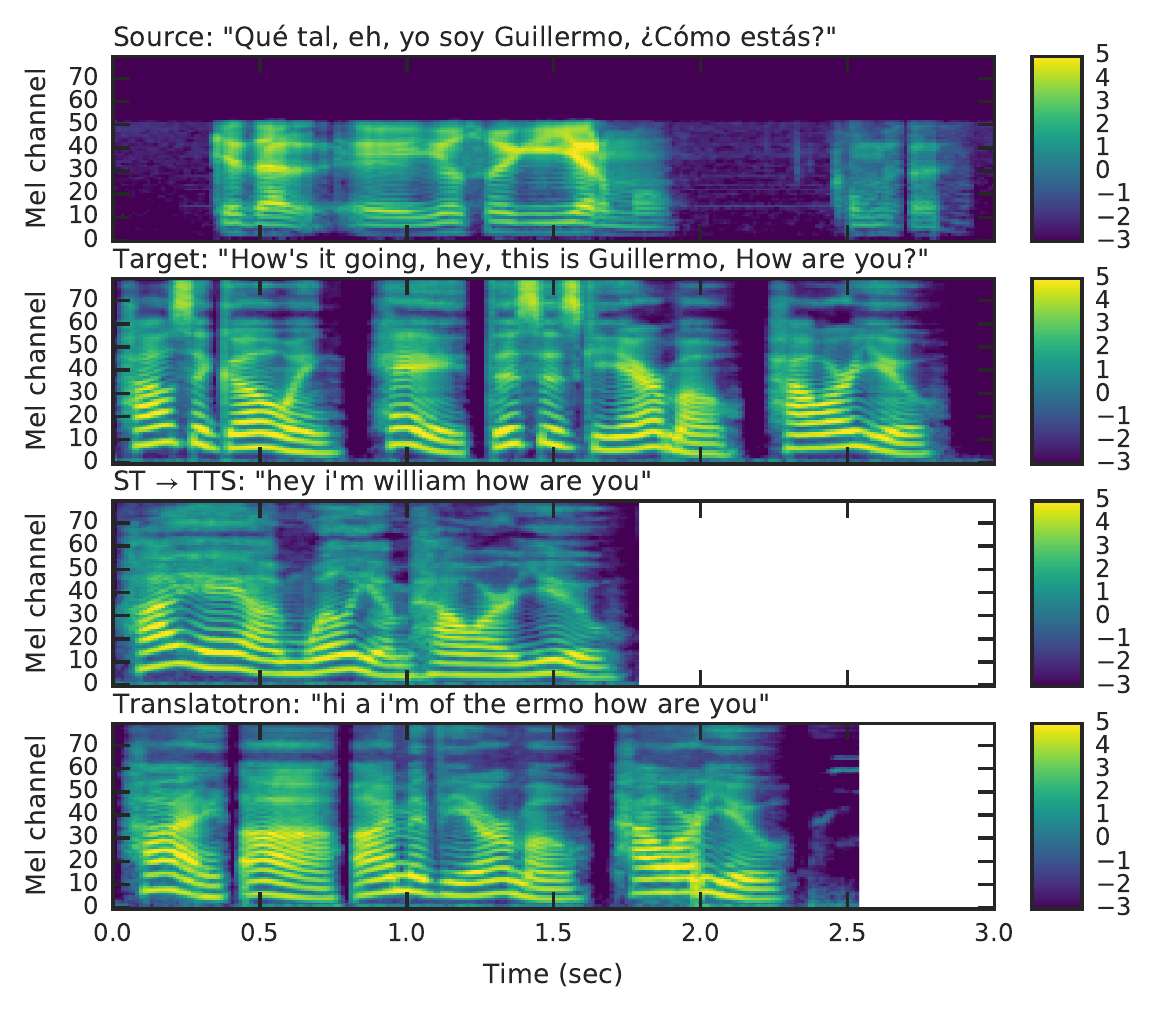}
   \vskip-2.75ex
  \caption{Mel spectrograms of input (top, upsampled to 24~kHz) and WaveRNN vocoder output (bottom) waveforms from a Fisher corpus example, along with ASR transcripts.
  Note that the spectrogram scales are different to the model inputs and outputs.
  Corresponding audio is on the companion website.
       }
  \vskip-2.6ex       
  \label{fig:example_outputs-fisher}
\end{figure}

Finally, we note the similar performance gap between \modelname{} and the baseline under this evaluation.
In part, this is a consequence of the different types of errors made by the two models.
For example, \modelname{} sometimes mispronounces words, especially proper nouns, using pronunciations from the source language, 
e.g.\ mispronouncing the /ae/ vowel in ``Dan'' as /ah/, consistent with Spanish but
sounding less natural to English listeners,
whereas 
by construction, 
the baseline consistently projects results to English.
Figure~\ref{fig:example_outputs-fisher} demonstrates other differences in behavior, where \modelname{} reproduces the input ``eh'' disfluency  (transcribed as ``a'', between $0.4-0.8$ sec in the bottom row of the figure), but the cascade does not. 
It is also interesting to note that the cascade translates ``Guillermo'' to its English form ``William'', whereas \modelname{} speaks the Spanish name (although the ASR model mistranscribes it as ``of the ermo''),
suggesting that the direct model might have a bias toward more directly reconstructing the input. Similarly, in example 7 on the companion page  \modelname{} reconstructs ``pasejo'' as ``passages'' instead of ``tickets'', potentially reflecting a bias for cognates.
We leave detailed analysis to future work.

\subsection{Cross language voice transfer}
\label{sec:voice_transfer}

In our final experiment, we
synthesize translated speech using the voice of the source speaker
by training the full model depicted in Figure~\ref{fig:model}. The speaker encoder is conditioned on the ground truth target speaker during training.
We use a subset of the data from Sec.~\ref{sec:proprietary} for which we have paired source and target recordings.
Note that the source and target speakers for each pair are always different -- the data was not collected from bilingual speakers.
This dataset contains 606k utterance pairs, resampled to 16~kHz, with 863 and 493~hours of source and target speech, respectively;
6.3k pairs, a subset of that from Sec.~\ref{sec:proprietary},
are held out for testing.
Since target recordings contained noise, we apply the denoising and volume normalization from \cite{jia2018transfer} to improve output quality.

\begin{table}[t]
\centering
\setlength{\tabcolsep}{0.25em}
\begin{small}
\caption{Voice transfer performance when conditioned on source, ground truth target, or a random utterance in the target language. References for MOS-similarity
match the conditioning speaker.
}
\vskip-1ex
\begin{tabular}{lccc}
    \toprule
    Speaker Emb & BLEU & MOS-naturalness & MOS-similarity \\
    \midrule
    Source        & 33.6 & $ 3.07 \pm 0.08$ & $ 1.85 \pm 0.06$ \\
    Target        & 36.2 & $ 3.15 \pm 0.08$ & $ 3.30 \pm 0.09$ \\
    Random target & 35.4 & $ 3.08 \pm 0.08$ & $ 3.24 \pm 0.08$ \\
    \midrule
    Ground truth  & 59.9 & $4.10 \pm 0.06$ & - \\
    \bottomrule
\end{tabular}
\label{tbl:voice-transfer}
\end{small}
\vskip-2.5ex
\end{table}

Table~\ref{tbl:voice-transfer} compares performance using different conditioning strategies.
The top row transfers the source speaker's voice to the translated speech, while row two is a ``cheating'' configuration since the speaker embedding can potentially leak information about the target content to the decoder.
To verify that this does not negatively impact performance
we also condition on random target utterances in row three.
In all cases performance is worse than models trained on synthetic targets in Tables~\ref{tbl:proprietary}~and~\ref{tbl:naturalness}. This is because  the task of synthesizing arbitrary speakers is more difficult;
 the training targets are much noisier and training set is much smaller;
 and
 the ASR model used for evaluation makes more errors on the noisy, multispeaker targets.
In terms of BLEU score,
the difference between conditioning on ground truth and random targets is very small, verifying
that content leak is not a concern (in part due to the low speaker embedding dimension).
However conditioning on the source trails by 1.8 BLEU points, reflecting the mismatch in conditioning language between the training and inference configurations.
Naturalness MOS scores are close in all cases.
However, conditioning on the source speaker significantly reduces similarity MOS by 1.4 points.
Again this suggests that using English speaker embeddings during training does not generalize well to Spanish speakers.

 \section{Conclusions}
\label{sec:conslusions}

We present a direct speech-to-speech translation model, trained end-to-end.
We find that it is important to use speech transcripts during training, but no
intermediate speech transcription is necessary for inference.
Exploring alternate training strategies which alleviate this requirement is an interesting direction for future work.
The model achieves high translation quality on two Spanish-to-English datasets, although performance is not as good as  a baseline cascade of ST and TTS models.

In addition, we demonstrate a variant which simultaneously transfers the source speaker's voice to the translated speech.
The voice transfer does not work as well as in a similar TTS context \cite{jia2018transfer}, reflecting the difficulty of the cross-language voice transfer task,
as well as evaluation \cite{wester2010speaker}.
Potential strategies to improve voice transfer performance include
improving the speaker encoder by adding a language adversarial loss, 
or by incorporating a cycle-consistency term \cite{nachmani2018fitting} into the S2ST loss.

Other future work includes utilizing weakly supervision to scale up training 
with synthetic data \cite{jia2018leveraging} or multitask learning \cite{weiss2017sequence, anastasopoulos2018tied}, and transferring prosody and other acoustic factors from the source speech to the translated speech following \cite{lee2018robust, wang2018style, hsu2018hierarchical}. 
\section{Acknowledgements}
The authors thank Quan Wang, Jason Pelecanos and the Google Speech team for providing the multilingual speaker encoder, Tom Walters and the Deepmind team for help with WaveNet TTS,
Quan Wang, Heiga Zen, Patrick Nguyen, Yu Zhang, Jonathan Shen, Orhan Firat, and the Google Brain team for helpful discussions,
and Mengmeng Niu for data collection support.

\bibliographystyle{IEEEtran}
\bibliography{references}

\begin{thebibliography}{10}
\providecommand{\url}[1]{#1}
\csname url@samestyle\endcsname
\providecommand{\newblock}{\relax}
\providecommand{\bibinfo}[2]{#2}
\providecommand{\BIBentrySTDinterwordspacing}{\spaceskip=0pt\relax}
\providecommand{\BIBentryALTinterwordstretchfactor}{4}
\providecommand{\BIBentryALTinterwordspacing}{\spaceskip=\fontdimen2\font plus
\BIBentryALTinterwordstretchfactor\fontdimen3\font minus
  \fontdimen4\font\relax}
\providecommand{\BIBforeignlanguage}[2]{{%
\expandafter\ifx\csname l@#1\endcsname\relax
\typeout{** WARNING: IEEEtran.bst: No hyphenation pattern has been}%
\typeout{** loaded for the language `#1'. Using the pattern for}%
\typeout{** the default language instead.}%
\else
\language=\csname l@#1\endcsname
\fi
#2}}
\providecommand{\BIBdecl}{\relax}
\BIBdecl

\bibitem{lavie1997janus}
A.~Lavie, A.~Waibel, L.~Levin, M.~Finke, D.~Gates, M.~Gavalda, T.~Zeppenfeld,
  and P.~Zhan, ``{JANUS-III}: Speech-to-speech translation in multiple
  languages,'' in \emph{Proc. ICASSP}, 1997.

\bibitem{wahlster2000verbmobil}
W.~Wahlster, \emph{Verbmobil: Foundations of speech-to-speech
  translation}.\hskip 1em plus 0.5em minus 0.4em\relax Springer, 2000.

\bibitem{nakamura2006atr}
S.~Nakamura, K.~Markov, H.~Nakaiwa, G.-i. Kikui, H.~Kawai, T.~Jitsuhiro, J.-S.
  Zhang, H.~Yamamoto, E.~Sumita, and S.~Yamamoto, ``The {ATR} multilingual
  speech-to-speech translation system,'' \emph{IEEE Transactions on Audio,
  Speech, and Language Processing}, 2006.

\bibitem{itu-f745}
{International Telecommunication Union}, ``{ITU-T} {F.745}: Functional
  requirements for network-based speech-to-speech translation services,'' 2016.

\bibitem{ney1999speech}
H.~Ney, ``Speech translation: Coupling of recognition and translation,'' in
  \emph{Proc. ICASSP}, 1999.

\bibitem{matusov2005integration}
E.~Matusov, S.~Kanthak, and H.~Ney, ``On the integration of speech recognition
  and statistical machine translation,'' in \emph{European Conference on Speech
  Communication and Technology}, 2005.

\bibitem{vidal1997finite}
E.~Vidal, ``Finite-state speech-to-speech translation,'' in \emph{Proc.
  ICASSP}, 1997.

\bibitem{casacuberta2004some}
F.~Casacuberta, H.~Ney, F.~J. Och, E.~Vidal, J.~M. Vilar \emph{et~al.}, ``Some
  approaches to statistical and finite-state speech-to-speech translation,''
  \emph{Computer Speech and Language}, vol.~18, no.~1, 2004.

\bibitem{aguero2006prosody}
P.~Aguero, J.~Adell, and A.~Bonafonte, ``Prosody generation for
  speech-to-speech translation,'' in \emph{Proc. ICASSP}, 2006.

\bibitem{do2017toward}
Q.~T. Do, S.~Sakti, and S.~Nakamura, ``Toward expressive speech translation: a
  unified sequence-to-sequence {LSTMs} approach for translating words and
  emphasis,'' in \emph{Proc. Interspeech}, 2017.

\bibitem{kano2018end}
T.~Kano, S.~Takamichi, S.~Sakti, G.~Neubig, T.~Toda, and S.~Nakamura, ``An
  end-to-end model for cross-lingual transformation of paralinguistic
  information,'' \emph{Machine Translation}, pp. 1--16, 2018.

\bibitem{kurimo2010personalising}
M.~Kurimo, W.~Byrne, J.~Dines, P.~N. Garner, M.~Gibson, Y.~Guan,
  T.~Hirsim{\"a}ki, R.~Karhila, S.~King, H.~Liang \emph{et~al.},
  ``Personalising speech-to-speech translation in the {EMIME} project,'' in
  \emph{Proc. ACL 2010 System Demonstrations}, 2010.

\bibitem{nachmani2018fitting}
E.~Nachmani, A.~Polyak, Y.~Taigman, and L.~Wolf, ``Fitting new speakers based
  on a short untranscribed sample,'' in \emph{ICML}, 2018.

\bibitem{arik2018neural}
S.~O. Arik, J.~Chen, K.~Peng, W.~Ping, and Y.~Zhou, ``Neural voice cloning with
  a few samples,'' in \emph{Proc. NeurIPS}, 2018.

\bibitem{jia2018transfer}
Y.~Jia, Y.~Zhang, R.~J. Weiss, Q.~Wang, J.~Shen, F.~Ren, Z.~Chen \emph{et~al.},
  ``Transfer learning from speaker verification to multispeaker text-to-speech
  synthesis,'' in \emph{Proc. NeurIPS}, 2018.

\bibitem{chen2018sample}
Y.~Chen, Y.~Assael, B.~Shillingford, D.~Budden, S.~Reed, H.~Zen, Q.~Wang, L.~C.
  Cobo, A.~Trask, B.~Laurie \emph{et~al.}, ``Sample efficient adaptive
  text-to-speech,'' in \emph{Proc. ICLR}, 2019.

\bibitem{berard2016listen}
A.~B{\'e}rard, O.~Pietquin, C.~Servan, and L.~Besacier, ``Listen and translate:
  A proof of concept for end-to-end speech-to-text translation,'' in
  \emph{NeurIPS Workshop on End-to-end Learning for Speech and Audio
  Processing}, 2016.

\bibitem{berard2018end}
A.~B{\'e}rard, L.~Besacier, A.~C. Kocabiyikoglu, and O.~Pietquin, ``End-to-end
  automatic speech translation of audiobooks,'' in \emph{Proc. ICASSP}, 2018.

\bibitem{weiss2017sequence}
R.~J. Weiss, J.~Chorowski, N.~Jaitly, Y.~Wu, and Z.~Chen,
  ``Sequence-to-sequence models can directly translate foreign speech,'' in
  \emph{Proc. Interspeech}, 2017.

\bibitem{anastasopoulos2018tied}
A.~Anastasopoulos and D.~Chiang, ``Tied multitask learning for neural speech
  translation,'' in \emph{Proc. NAACL-HLT}, 2018.

\bibitem{jia2018leveraging}
Y.~Jia, M.~Johnson, W.~Macherey, R.~J. Weiss, Y.~Cao, C.-C. Chiu, N.~Ari
  \emph{et~al.}, ``Leveraging weakly supervised data to improve end-to-end
  speech-to-text translation,'' in \emph{Proc. ICASSP}, 2019.

\bibitem{haque2018conditional}
A.~Haque, M.~Guo, and P.~Verma, ``Conditional end-to-end audio transforms,'' in
  \emph{Proc. Interspeech}, 2018.

\bibitem{zhang2019sequence}
J.~Zhang, Z.~Ling, L.-J. Liu, Y.~Jiang, and L.-R. Dai, ``Sequence-to-sequence
  acoustic modeling for voice conversion,'' \emph{IEEE/ACM Transactions on
  Audio, Speech, and Language Processing}, 2019.

\bibitem{biadsy2019parrotron}
F.~Biadsy, R.~J. Weiss, P.~J. Moreno, D.~Kanevsky, and Y.~Jia, ``Parrotron: An
  end-to-end speech-to-speech conversion model and its applications to
  hearing-impaired speech and speech separation,'' in \emph{Proc. Interspeech},
  2019.

\bibitem{guo2019end}
M.~Guo, A.~Haque, and P.~Verma, ``End-to-end spoken language translation,''
  \emph{arXiv preprint arXiv:1904.10760}, 2019.

\bibitem{shen2018natural}
J.~Shen, R.~Pang, R.~J. Weiss, M.~Schuster, N.~Jaitly, Z.~Yang, Z.~Chen,
  Y.~Zhang \emph{et~al.}, ``Natural {TTS} synthesis by conditioning {WaveNet}
  on mel spectrogram predictions,'' in \emph{Proc. ICASSP}, 2017.

\bibitem{machado2010voice}
A.~F. Machado and M.~Queiroz, ``Voice conversion: A critical survey,'' in
  \emph{Proc. Sound and Music Computing}, 2010, pp. 1--8.

\bibitem{chiu2017state}
C.-C. Chiu, T.~Sainath, Y.~Wu, R.~Prabhavalkar, P.~Nguyen, Z.~Chen, A.~Kannan,
  R.~Weiss, K.~Rao \emph{et~al.}, ``State-of-the-art speech recognition with
  sequence-to-sequence models,'' in \emph{Proc. ICASSP}, 2018.

\bibitem{vaswani2017attention}
A.~Vaswani, N.~Shazeer, N.~Parmar, J.~Uszkoreit, L.~Jones, A.~N. Gomez,
  {\L}.~Kaiser, and I.~Polosukhin, ``Attention is all you need,'' in
  \emph{Proc. NeurIPS}, 2017.

\bibitem{yx2017tacotron}
Y.~Wang, R.~Skerry-Ryan, D.~Stanton, Y.~Wu, R.~J. Weiss, N.~Jaitly, Z.~Yang,
  Y.~Xiao, Z.~Chen, S.~Bengio, Q.~Le \emph{et~al.}, ``Tacotron: Towards
  end-to-end speech synthesis,'' in \emph{Proc. Interspeech}, 2017.

\bibitem{wu2016google}
Y.~Wu, M.~Schuster, Z.~Chen, Q.~V. Le, M.~Norouzi, W.~Macherey, M.~Krikun,
  Y.~Cao, Q.~Gao, K.~Macherey \emph{et~al.}, ``Google's neural machine
  translation system: Bridging the gap between human and machine translation,''
  \emph{arXiv:1609.08144}, 2016.

\bibitem{bahdanau2014neural}
D.~Bahdanau, K.~Cho, and Y.~Bengio, ``Neural machine translation by jointly
  learning to align and translate,'' in \emph{Proc. ICLR}, 2015.

\bibitem{krueger2016zoneout}
D.~Krueger, T.~Maharaj, J.~Kram{\'a}r, M.~Pezeshki, N.~Ballas, N.~R. Ke,
  A.~Goyal, Y.~Bengio \emph{et~al.}, ``Zoneout: Regularizing {RNNs} by randomly
  preserving hidden activations,'' in \emph{Proc. {ICLR}}, 2017.

\bibitem{shazeer2018adafactor}
N.~Shazeer and M.~Stern, ``Adafactor: Adaptive learning rates with sublinear
  memory cost,'' in \emph{Proc. ICML}, 2018, pp. 4603--4611.

\bibitem{griffin1984signal}
D.~Griffin and J.~Lim, ``Signal estimation from modified short-time {F}ourier
  transform,'' \emph{IEEE Transactions on Acoustics, Speech, and Signal
  Processing}, vol.~32, no.~2, pp. 236--243, 1984.

\bibitem{kalchbrenner2018efficient}
N.~Kalchbrenner, E.~Elsen, K.~Simonyan, S.~Noury, N.~Casagrande, E.~Lockhart,
  F.~Stimberg, A.~v.~d. Oord, S.~Dieleman \emph{et~al.}, ``Efficient neural
  audio synthesis,'' in \emph{Proc. ICML}, 2018.

\bibitem{zhang2018fully}
A.~Zhang, Q.~Wang, Z.~Zhu, J.~Paisley, and C.~Wang, ``Fully supervised speaker
  diarization,'' in \emph{Proc. ICASSP}, 2019.

\bibitem{post2013improved}
M.~Post, G.~Kumar, A.~Lopez, D.~Karakos, C.~Callison-Burch \emph{et~al.},
  ``Improved speech-to-text translation with the {Fisher and Callhome
  Spanish--English} speech translation corpus,'' in \emph{Proc. IWSLT}, 2013.

\bibitem{shen2019lingvo}
J.~Shen, P.~Nguyen, Y.~Wu, Z.~Chen \emph{et~al.}, ``Lingvo: a modular and
  scalable framework for sequence-to-sequence modeling,'' 2019.

\bibitem{papineni-EtAl:2002:ACL}
K.~Papineni, S.~Roukos, T.~Ward, and W.-J. Zhu, ``{BLEU}: A method for
  automatic evaluation of machine translation,'' in \emph{ACL}, 2002.

\bibitem{irie2019model}
K.~Irie, R.~Prabhavalkar, A.~Kannan, A.~Bruguier, D.~Rybach, and P.~Nguyen,
  ``Model unit exploration for sequence-to-sequence speech recognition,''
  \emph{arXiv:1902.01955}, 2019.

\bibitem{panayotov2015librispeech}
V.~Panayotov, G.~Chen, D.~Povey, and S.~Khudanpur, ``{LibriSpeech}: an {ASR}
  corpus based on public domain audio books,'' in \emph{Proc. ICASSP}, 2015.

\bibitem{pmlr-v80-oord18a}
A.~van~den Oord, Y.~Li, I.~Babuschkin, K.~Simonyan, O.~Vinyals, K.~Kavukcuoglu,
  G.~van~den Driessche \emph{et~al.}, ``Parallel {W}ave{N}et: Fast
  high-fidelity speech synthesis,'' in \emph{Proc. ICML}, 2018.

\bibitem{wester2010speaker}
M.~Wester, J.~Dines, M.~Gibson, H.~Liang \emph{et~al.}, ``Speaker adaptation
  and the evaluation of speaker similarity in the {EMIME} speech-to-speech
  translation project,'' in \emph{ISCA Tutorial and Research Workshop on Speech
  Synthesis}, 2010.

\bibitem{lee2018robust}
Y.~Lee and T.~Kim, ``Robust and fine-grained prosody control of end-to-end
  speech synthesis,'' in \emph{Proc. ICASSP}, 2019.

\bibitem{wang2018style}
Y.~Wang, D.~Stanton, Y.~Zhang, R.~Skerry-Ryan, E.~Battenberg, J.~Shor
  \emph{et~al.}, ``Style tokens: Unsupervised style modeling, control and
  transfer in end-to-end speech synthesis,'' in \emph{Proc. ICML}, 2018.

\bibitem{hsu2018hierarchical}
W.-N. Hsu, Y.~Zhang, R.~J. Weiss, H.~Zen, Y.~Wu, Y.~Wang, Y.~Cao, Y.~Jia,
  Z.~Chen, J.~Shen \emph{et~al.}, ``Hierarchical generative modeling for
  controllable speech synthesis,'' in \emph{Proc. ICLR}, 2019.

\end{thebibliography}

\end{document}